\documentclass[11pt,letterpaper]{article}
\usepackage{naaclhlt2015}
\usepackage{times}
\usepackage{latexsym}
\usepackage{url}
\usepackage{graphicx}
\title{Discriminative Phrase Embedding for Paraphrase Identification}

\author{Wenpeng Yin \rm{and} \textbf{Hinrich Sch\"utze}\\
Center for Information and Language Processing\\
University of Munich, Germany\\
{\tt wenpeng@cis.lmu.de}} 

\date{}

\newcommand{\bftab}{\fontseries{b}\selectfont}

\newcommand{\enotesoff}{\long\gdef\enote##1##2{}}
\newcommand{\enoteson}{\long\gdef\enote##1##2{{\large\bf
\hspace{1cm}$<<<$ ##1: ##2 $>>>$\hspace{1cm}}}}
\enoteson
\enotesoff

\def\secref#1{Section~\ref{sec:#1}}
\def\seclabel#1{\label{sec:#1}\label{p:#1}}
\def\eqref#1{Eq.~\ref{eqn:#1}}

\begin{document}
\maketitle
\begin{abstract}
This work, concerning paraphrase identification task, on one
hand contributes to expanding deep learning embeddings to
include continuous and discontinuous linguistic phrases. On
the other hand, it comes up with a new scheme TF-KLD-KNN to
learn the discriminative weights of words and phrases specific to
paraphrase task, so that a weighted sum of embeddings can
represent sentences more effectively.  Based on these two
innovations we get competitive state-of-the-art performance on
paraphrase identification.
\end{abstract}

\section{Introduction}\seclabel{intro}

This work investigates representation learning via deep
learning in paraphrase identification task, which aims to
determine whether two sentences have the same meaning. One
main innovation of deep learning is that it learns distributed word
representations (also called ``word embeddings'') to deal with various Natural Language Processing
(NLP) tasks. Our goal is to use and refine embeddings to get
competitive performance.

We adopt a supervised classification approach to paraphrase
identification like
most top performing systems.
Our focus is 
representation learning of sentences. 
Following prior work (e.g., \newcite{blacoe2012comparison}), we compute the
vector of a sentence as the sum of the vectors of its
components.
But unlike prior work
we use \emph{single words, continuous phrases and
discontinuous phrases} as the components, not just single words.
Our rationale is that 
many semantic
units are formed by multiple words --  e.g., the continuous phrase
``side effects'' and the discontinuous phrase
``pick \ldots\ off''.
The better we can discover and
represent such components, the better the compositional
sentence vector should be. 
We use the term \emph{unit} to refer to 
single words, continuous phrases and
discontinuous phrases.

\newcite{jidiscriminative} show that 
not all words are equally important for paraphrase
identification.
They propose TF-KLD, a
discriminative weighting scheme to address this problem.
While they do not represent sentences as vectors composed of
other vectors, TF-KLD is promising for a vector-based
approach as well since the insight that units are of
different importance still applies.
A shortcoming of TF-KLD is its failure to define weights for
words that do not occur in the training set.
We propose TF-KLD-KNN, an
extension of TF-KLD that computes
the weight of an unknown unit as the
average of the weights of its $k$ nearest neighbors.  We
determine nearest neighbors by cosine measure over embedding
space.
We then represent a sentence as the sum of the vectors of
its units, weighted by TF-KLD-KNN.

We use \cite{madnani2012re} as our baseline system. They
used simple features -- eight different machine translation
metrics -- yet got good performance. Based on above new
sentence representations, we compute three kinds of features
to describe a pair of sentences -- cosine similarity,
element-wise sum and absolute element-wise difference -- and show
that combining them with the features from
\newcite{madnani2012re} gets state-of-the-art performance on
the Microsoft Research Paraphrase (MSRP) corpus
\cite{dolan2004unsupervised}.

In summary, our first contribution lies in
embedding learning of continuous and discontinuous
phrases. Our second contribution is the weighting scheme TF-KLD-KNN.

This paper is structured as follows.
\secref{related} reviews related work.
\secref{learning} describes our method
for learning embeddings of units.
\secref{measure} introduces
a measure of unit discriminativity that can be used for
differential weighting of units.
\secref{experiments} presents experimental setup and results.
\secref{conclusion} concludes.

\section{Related work}
\seclabel{related}
The key for good performance in paraphrase identification is
the design of good features.
We now discuss relevant prior work based on the
linguistic
granularity of feature learning.

The first line is compositional semantics, which learns
representations for words and then composes
them to representations of sentences. \newcite{blacoe2012comparison} carried out a
comparative study of three word representation methods (the
simple distributional semantic space
\cite{mitchell2010composition}, distributional memory tensor
\cite{baroni2010distributional} and word embedding
\cite{collobert2008unified}), along with three composition
methods (addition, point-wise multiplication, and recursive
auto-encoder \cite{socher2011dynamic}). They showed that addition
over word embeddings is competitive, despite  its
simplicity.

The second category directly seeks sentence-level
features. \newcite{jidiscriminative} explored
unigrams, bigrams and dependency pairs as sentence
features. They proposed TF-KLD to weight  features and
used non-negative factorization to
learn latent sentence representations. 
Our method TF-KLD-KNN is an extension of their work.

The third line directly computes features for sentence
pairs. \newcite{wan2006using} used N-gram
overlap, dependency relation overlap, dependency tree-edit
distance and difference of sentence lengths. 
\newcite{finch2005using} and \newcite{madnani2012re} combined several
machine translation metrics. 
\newcite{das2009paraphrase} presented a generative model over two sentences' dependency trees, incorporating syntax, lexical semantics, and hidden loose alignments between the trees to model generating a paraphrase of a given sentence. \newcite{socher2011dynamic} used recursive
autoencoders to learn representations for words and word
sequences
on each layer of the sentence parsing tree, and then proposed
dynamic pooling layer to form a fixed-size matrix as the
representation of the two sentences. Other 
work 
representative of this line 
is by
\newcite{kozareva2006paraphrase},
\newcite{qiu2006paraphrase},
\newcite{zia2012}.

Our work, first learning unit embeddings, then adding them to
form sentence representations, finally calculating pair
features (cosine similarity, absolute difference and MT metrics) actually is a combination of
above three lines.


\section{Embedding learning for units}
\seclabel{learning}
As explained in \secref{intro}, ``units'' in this work
include single words, continuous phrases and discontinuous
phrases. Phrases have a larger linguistic granularity than
words and thus will in general
contain more meaning aspects for a sentence. For
example, successful detection of continuous phrase ``side
effects'' and discontinuous phrase ``pick $\cdots$ off'' is
helpful to understand the sentence meaning correctly.
This section focuses on how to
detect phrases and how to represent them.

\subsection{Phrase collection}
Phrases defined by a lexicon have not been investigated
extensively before
in deep learning. To collect canonical phrase set, we
extract two-word phrases defined in
Wiktionary\footnote{\url{http://en.wiktionary.org}} and Wordnet
\cite{miller1998wordnet} to form a collection of size 
95,218. 
This collection contains \emph{continuous phrases} --
phrases whose parts
always occur next to each other (e.g., ``side effects'') -- and
\emph{discontinuous phrases} -- phrases  whose parts more often occur separated from each
other (e.g., ``pick \ldots\ off''). 

\subsection{Identification of phrase continuity}
Wiktionary and WordNet do not categorize phrases as
continuous or discontinuous. So we need a heuristic to
determine this automatically.

For each phrase ``A\_B'', we compute [$c_1$, $c_2$, $c_3$, $c_4$,
  $c_5$] where $c_i, 1\leq i\leq 5$, indicates there are
$c_i$ occurrences of A and B in that order with a distance
of $i$. We compute these statistics for a corpus consisting
of English Gigaword \cite{graff2003english} and Wikipedia.  We set the maximal distance to 5
because discontinuous phrases are rarely
separated by more than 5 tokens.

If $c_1$ is 10 times higher than $(c_2+c_3+c_4+c_5)/4$, we
classify ``A\_B'' as \emph{continuous}, otherwise as
\emph{discontinuous}.
For example, 
[$c_1$, \ldots,
$c_5$] 
is [1121, 632, 337, 348, 4052] for
``pick\_off'', so
$c_1$
is smaller than the average 1342.25 and  ``pick\_off''
is set as ``discontinuous''; 
[$c_1$, \ldots,
$c_5$] 
is 
[14831, 16, 177, 331, 3471] for
``Cornell
University'', $c_1$ is 10 times larger than the average and this
phrase is set to 
``continuous''.

We found that that this heuristic for distinguishing
between continuous and discontinuous phrases works well and
leave the development of a more principled method for future work.

\subsection{Sentence reformatting}
Sentence ``\ldots\ A \ldots\ B \ldots'' is
\begin{itemize} 
\item reformatted as ``\ldots\ A\_B \ldots'' if A and B form a continuous
phrase and no word intervenes between them and
\item reformatted  as
``\ldots\ A\_B \ldots\ A\_B \ldots'' if A and B form a
discontinuous phrase and are separated by 1 to  4 words.
We replace each of
the two component words 
with A\_B to
make the context of
both constituents available to the phrase in learning. 
\end{itemize}

This method of phrase detection will generate some false
positives, e.g., if ``pick'' and ``off'' occur
in a context like ``she picked an island off the coast of
Maine''. However, our experimental results indicate that it
is robust enough for our purposes.

We run
word2vec
\cite{mikolov2013distributed}  on the reformatted Wikipedia corpus
to learn embeddings for all units.  
Embedding
size is set to 200.

\section{Measure of unit discriminativity}
\seclabel{measure}
We will represent a sentence as the sum of the embeddings of
its units.  Building on  \newcite{jidiscriminative}'s TF-KLD, we want
to weight units according to their ability to discriminate
two sentences specific to the paraphrase task.

TF-KLD assumes a training set of sentence pairs in the form
$\left\langle u_i, v_i, t_i\right\rangle$, where
$u_i$ and $v_i$ denote the binary unit occurrence vectors for the
sentences in the $i$th pair  and $t_i \in \{0,1\}$ is the gold tag. Then,
we define $p_k$ and $q_k$ as follows.
\begin{itemize}
\itemsep=0.01cm
\item $p_k=P(u_{ik}|v_{ik}=1, t_i=1)$. This is the
  probability that 
unit $w_k$ occurs in 
sentence $u_i$  given that $w_k$ occurs in 
its counterpart $v_i$  and they are paraphrases.
\item $q_k=P(u_{ik}|v_{ik}=1, t_i=0)$. This is the
  probability that 
unit $w_k$ occurs in 
sentence $u_i$  given that $w_k$ occurs in 
its counterpart $v_i$  and they are not paraphrases.
\end{itemize} 

TF-KLD computes the discriminativity of unit $w_k$ as the
Kullback-Leibler divergence of the Bernoulli distributions ($p_k$, 1-$p_k$) and ($q_k$, 1-$q_k$)

TF-KLD has a serious shortcoming for unknown units.
Unfortunately, the test data of the commonly used MSPR
corpus in paraphrase task has about 6\% unknown words and 62.5\%
of its sentences contain unknown words. It motivates us to
design an improved scheme TF-KLD-KNN to reweight the
features.

TF-KLD-KNN weights are the same as TF-KLD weights for known
units.  For a unit that did not occur in training,
TF-KLD-KNN computes its weight as the average of the weights
of its $k$ nearest neighbors in embedding space, where unit
similarity is calculated by cosine measure.\footnote{Unknown
  words without embeddings 
 (only seven cases in our
  experiments)
are ignored. This problem can be effectively relieved by
  training embedding on larger corpora.}

Word2vec learns word embeddings based on
the word context. 
The intuition of TF-KLD-KNN is that 
words with similar
context have similar discriminativities. This enables us to
transfer the weights of features in training data to the
unknown features in test data, greatly helping 
to address problems of sparseness.

\section{Experiments}
\seclabel{experiments}
\subsection{Data and baselines}
We use the MSRP corpus
\cite{dolan2004unsupervised} for evaluation.  It consists of
a training set of 2753 true paraphrase pairs and 1323
false paraphrase pairs and a test set of 1147 true
and 578 false pairs. 

For our new method,
it is interesting to measure
the
improvement on 
the subset of those MSRP sentences that contain at least one phrase.
In the standard MSRP corpus, 3027 training pairs
(2123 true, 904 false) and 1273 test pairs (871 true,
402 false) contain phrases; we denote
this subset as \emph{subset}. We carry out experiments on
\emph{overall} (all MSRP sentences) as well as \emph{subset} cases.  

We compare six methods for paraphrase identification.

\begin{itemize}

\item \textbf{NOWEIGHT.} Following \newcite{blacoe2012comparison},
we simply represent a sentence as the unweighted sum of
the embeddings of all its units.

\item \textbf{MT}  is the method proposed by \newcite{madnani2012re}: the
sentence pair is represented as a vector of eight
different machine translation metrics. 

\item \textbf{\newcite{jidiscriminative}}. We reimplemented their ``inductive'' setup which is based on matrix factorization and is the top-performing system in paraphrasing task.\footnote{They report even better performance
in a 
``transductive'' setup that makes use of test data. We only
address paraphrase identification for the case that the test
data are not available for training the model in this paper.}

The following three methods not only use this vector of eight
MT metrics, but use three kinds of additional features given two sentence representations $s_1$ and $s_2$: cosine similarity, element-wise sum $s_1+s_2$ and element-wise absolute difference $|s_1-s_2|$. We now describe how each of the three
methods computes the sentence vectors.

\item \textbf{WORD.} The sentence is represented as the sum of all
single-word embeddings, weighted by TF-KLD-KNN.

\item \textbf{WORD+PHRASE.} The sentence is represented as the sum of the
embeddings of all its units (including phrases), weighted by
TF-KLD-KNN.

\item \textbf{WORD+GOOGLE.} \newcite{mikolov2013distributed} use
a data-driven method to detect statistical phrases
which are mostly continuous bigrams. We implement their
system by first exploiting word2phrase\footnote{\url{https://code.google.com/p/word2vec/}}
to reformat 
Wikipedia, then using word2vec skip-gram model to train
phrase embeddings. 

\end{itemize}

We use the same weighting scheme TF-KLD-KNN for the three
weighted sum approaches: 
WORD,
WORD+PHRASE and
WORD+GOOGLE. Note however that
there is an interaction between representation space and
nearest neighbor search.
We limit the neighbor range of unknown words for WORD to
single words; in contrast, we search the space of all single
words and linguistic (resp.\ Google) phrases for
WORD+PHRASE (resp.\ WORD+GOOGLE).

We use LIBLINEAR  \cite{fan2008liblinear} as our linear SVM implementation. 20\% training data is used as development data. Parameter $k$ is fine-tuned on development set and the best value 3 is finally used in following reported results.
\subsection{Experimental results}
Table \ref{tab:overall} shows performance for the six
methods as well as for the majority baseline. In the
\emph{overall} (resp.\ \emph{subset}) setup, WORD+PHRASE
performs best and outperforms \cite{jidiscriminative} by
.009 (resp.\ .052) on accuracy. Interestingly,
\newcite{jidiscriminative}'s method obtains worse performance on
$subset$. 
This can be explained by
the effect of matrix
factorization in their work: 
it works less well for smaller datasets like $subset$.
This is a shortcoming
of their approach. WORD+GOOGLE has a slightly worse
performance than WORD+PHRASE; this suggests that linguistic phrases might be more
effective than statistical phrases in identifying
paraphrases.

\begin{table}[tb]
 \setlength\tabcolsep{3pt}
\begin{center}
\begin{tabular}{l|ll| ll}
 &  \multicolumn{2}{c|}{overall} &  \multicolumn{2}{c}{subset}\\
method & acc & $F_1$ & acc & $F_1$\\ \hline
baseline & .665 & .799 & .684 & .812\\ 
NOWEIGHT & .708& .809 & .713 & .823\\
MT & .774& .841 & .772 & .839\\
\newcite{jidiscriminative} & .778 & .843 & .749 & .827 \\
WORD & .775 & .839 & .776 & .843\\
WORD+GOOGLE & .780 & .843 & .795 & .853\\
WORD+PHRASE & \bftab{.787} & \bftab{.848}$^*$ & \bftab{.801} & \bftab{.857}$^*$
\end{tabular}
\end{center}
\caption{Results on overall and subset corpus. Significant
  improvements over MT are marked with $*$ (approximate
  randomization test, \newcite{sigf06}, $p<.05$).}\label{tab:overall} 
\end{table}

Cases \emph{overall} and \emph{subset} both suggest 
that phrase embeddings improve
sentence representations. The accuracy of WORD+PHRASE is
lower on \emph{overall} than on
\emph{subset} because 
WORD+PHRASE has no advantage over WORD for sentences without
phrases.

\subsection{Effectiveness of TF-KLD-KNN}
The key contribution
of TF-KLD-KNN is that it achieves full coverage of feature weights 
in the face of data sparseness.
We now  compare four
weighting methods
on overall corpus and with the combination of MT features:
NOWEIGHT, TF-IDF, TF-KLD, TF-KLD
\begin{table}[tb]
\begin{center}
\begin{tabular}{l|ll}
method &  acc & \textbf{$ F_1$} \\ \hline
NOWEIGHT & .746  & .815\\ 
TF-IDF & .752& .821\\
TF-KLD & .774& .842\\
TF-KLD-KNN & \bftab{.787} & \bftab{.848}
\end{tabular}
\end{center}
\caption{Effects of different reweighting methods on overall.}\label{tab:reweight} 
\end{table}

Table \ref{tab:reweight} suggests that task-specific
reweighting approaches (including TF-KLD and TF-KLD-KNN) are superior to
unspecific schemes (NOWEIGHT and TF-IDF). Also, it
demonstrates the effectiveness of our weight learning
solution for unknown units in paraphrase task.

\subsection{Reweighting schemes for unseen units}
\begin{figure}[htbp]
\centering
\includegraphics[width=0.5\textwidth]{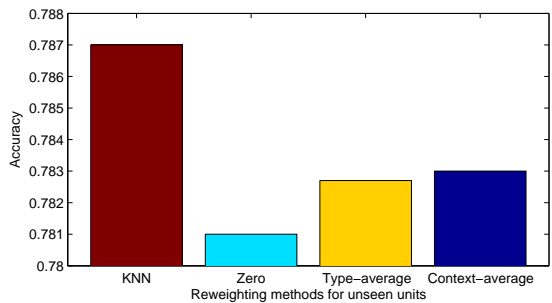}
\caption{Performance of different reweighting schemes for
  unseen units on overall.} \label{fig:reweightunseen}
\end{figure}

We compare our reweighting scheme \textbf{KNN} (i.e.,
TF-KLD-KNN) with three
other reweighting schemes.  \textbf{Zero}: zero weight,
i.e., ignore unseen units; \textbf{Type-average}: take the
average of weights of all known unit types in test set;
\textbf{Context-average}: average of the weights of the
adjacent known units of the unknown unit (two, one or
defaulting to Zero, depending on how many there are).
Figure \ref{fig:reweightunseen} shows that KNN performs
best.

\section{Conclusion}
\seclabel{conclusion}
This work introduced TF-KLD-KNN,
a new reweighting scheme
that learns the discriminativities of known as
well as unknown units  effectively. 
We further improved paraphrase identification performance by the utilization of continuous and
discontinuous phrase embeddings.

In future, we plan to do
experiments in a cross-domain setup and enhance our
algorithm for domain adaptation paraphrase identification.

\section*{Acknowledgments}
We are grateful to members of CIS
for comments on earlier
versions of this paper. This work was supported by
Baidu (through a Baidu scholarship awarded to Wenpeng Yin)
and by Deutsche Forschungsgemeinschaft (grant DFG SCHU 2246/8-2,
SPP 1335).


\bibliographystyle{naaclhlt2015}
\bibliography{acl}

\end{document}